\documentclass[times,twocolumn,final]{elsarticle}

\usepackage{appendix}
\usepackage{medima}
\usepackage{framed,multirow}

\usepackage{amssymb}
\usepackage{latexsym}

\usepackage{url}
\usepackage{xcolor}

\usepackage{hyperref}

\usepackage{adjustbox}

\usepackage{graphicx}
\usepackage{amsmath,amssymb}
\usepackage{xspace}

\usepackage{multirow,setspace,verbatim,amsfonts,amsmath,amsbsy,epsfig,url}
\usepackage{gensymb}
\usepackage{booktabs}
\usepackage{array,multirow}
\newcolumntype{P}[1]{>{\centering\arraybackslash}p{X1}}
\newcolumntype{M}[1]{>{\centering\arraybackslash}m{X1}}
\newcolumntype{Q}[1]{>{\arraybackslash}m{X1}}
\usepackage[flushleft]{threeparttable} 

\usepackage{algorithm}
\usepackage{lineno}
\usepackage[noend]{algpseudocode}
\usepackage{tikz}
\usepackage{mathrsfs} 
\usepackage[algo2e, linesnumbered]{algorithm2e} 
\usepackage{array}
\usepackage{color}
\usepackage{tabularx}
\usepackage{makecell}
\usepackage{bm}

\usepackage{pifont}
\usepackage{lipsum,afterpage,refcount}

\newcommand{\add}[1] {\textcolor{black}{#1}} 

\definecolor{newcolor}{rgb}{.8,.349,.1}

\journal{Medical Image Analysis}

\begin{document}

\verso{S Park and IJ Lee \textit{et~al.}}

\begin{frontmatter}

\title{Objective and Interpretable Breast Cosmesis Evaluation with Attention Guided Denoising Diffusion Anomaly Detection Model}


\author[1]{Sangjoon \snm{Park}}
\author[1]{Yong Bae \snm{Kim}}
\author[1]{Jee Suk \snm{Chang}}
\author[1]{Seo Hee \snm{Choi}}
\author[2]{Hyungjin \snm{Chung}}
\author[1]{Ik Jae \snm{Lee}\corref{cor2}\fnref{fn1}}
\fntext[fn1]{Co-corresponding authors.}
\cortext[cor2]{Corresponding author: 
  Tel.: +82-2-2228-8117;  
  fax: +82-8-2227-7823;}
  \ead{IKJAE412@yuhs.ac}
\author[3]{Hwa Kyung \snm{Byun}\corref{cor1}\fnref{fn1}}
\cortext[cor1]{Corresponding author: 
  Tel.: +82-31-5189-8166;  
  \add{fax: +82-8-2227-7823;}}
\ead{HKBYUN05@yuhs.ac}

\address[1]{Department of Radiation Oncology, Yonsei College of Medicine, Seoul, Korea}
\address[2]{Department of Bio and Brain Engineering, Korea Advanced Institute of Science and Technology, Daejeon, Korea}
\address[3]{Department of Radiation Oncology, Yongin Severance Hospital, Yonsei University College of Medicine, Yongin, Gyeonggi-do, Korea}

\received{}
\finalform{}
\accepted{}
\availableonline{}
\communicated{}

\begin{abstract}

As advancements in the field of breast cancer treatment continue to progress, the assessment of post-surgical cosmetic outcomes has gained increasing significance due to its substantial impact on patients' quality of life. However, evaluating breast cosmesis presents challenges due to the inherently subjective nature of expert labeling. In this study, we present a novel automated approach, \add{Attention-Guided Denoising Diffusion Anomaly Detection (AG-DDAD)}, designed to assess breast cosmesis following surgery, addressing the limitations of conventional supervised learning and existing anomaly detection models. Our approach leverages the attention mechanism of the \add{distillation with no label (DINO)} self-supervised Vision Transformer (ViT) in combination with a diffusion model to achieve high-quality image reconstruction and precise transformation of discriminative regions. By training the diffusion model on unlabeled data predominantly with normal cosmesis, we adopt an unsupervised anomaly detection perspective to automatically score the cosmesis. Real-world data experiments demonstrate the effectiveness of our method, providing visually appealing representations and quantifiable scores for cosmesis evaluation. Compared to commonly used rule-based programs, our fully automated approach eliminates the need for manual annotations and offers objective evaluation. Moreover, our anomaly detection model exhibits state-of-the-art performance, surpassing existing models in accuracy. 
Going beyond the scope of breast cosmesis, our research represents a significant advancement in unsupervised anomaly detection within the medical domain, thereby paving the way for future investigations.

\end{abstract}


\end{frontmatter}


\section{Introduction}
\label{sec1}

\add{The cosmetic outcome after breast cancer surgery carries considerable significance in relation to the quality of life experienced by patients \citep{sayan2017long, park2012unmet}. It is recognized as a crucial aspect impacting patients' well-being, considering breast cancer's status as the most prevalent cancer affecting women, with approximately 2.26 million new cases reported globally in 2020 \citep{arnold2022current}.}
Individuals who have undergone breast cancer surgery often express concerns regarding the shape, size, and overall appearance of their breasts. These concerns can exert a detrimental impact on their psychological and social well-being \citep{kim2015effect, volders2017cosmetic, leser2021patient}.
Various methods exist for evaluating the cosmetic outcome of breast cancer surgery. The modified Harvard-Harris cosmetic scale is a commonly employed subjective scoring system used to assess the cosmetic outcome following breast-conserving surgery \citep{zellars2009feasibility}. This scale focuses on aspects such as symmetry, general appearance, shape, size, color, and nipple position of both breasts, categorizing the cosmesis into four groups: excellent, good, fair, and poor.
However, the modified Harvard-Harris cosmetic scale is subject to subjective interpretation and suffers from low inter-rater reliability. To overcome these limitations, BCCT.core, a software tool, has been developed for the objective evaluation of cosmetic outcomes in BCS using digital photographs \citep{preuss2012bcct}. BCCT.core automatically measures various features in the photographs, including symmetry, shape, size, consistency, color, and nipple position. These measured features are then utilized to calculate a single score through a rule-based approach.

In recent years, there has been a growing interest in the utilization of deep learning-based artificial intelligence (AI) models for the assessment of cosmetic outcomes in breast conserving surgery. \cite{guo2022fully} demonstrated the potential of DL algorithms in evaluating breast cosmesis by accurately detecting and delineating the breast contour using expert-labeled data. However, this approach is inherently limited by the constraints of traditional supervised learning-based DL algorithms.
Firstly, these approaches heavily rely on the availability of data and the accuracy of ground truth labels for effective training. The dependence on expert annotation necessitates a labor-intensive process of individually annotating images, incurring significant costs and impeding the construction of large-scale datasets. Furthermore, models trained using this approach are constrained by the performance limitations of the ground truth labels, which may contain inherent inaccuracies and possibilities of inherent biases, especially in cases where subjective evaluation is involved during the labeling process, as in the assessment of breast cosmesis. Additionally, traditional DL methods are susceptible to overfitting issues when confronted with limited data, hindering their ability to generalize well across diverse scenarios \citep{brigato2021close}. Lastly, conventional supervised DL models often lack interpretability, as they have limited capacity to generate high-quality visualizations for comprehensive understanding.
In contrast, \cite{kim2022feasibility} explored the feasibility of utilizing an anomaly score for breast cancer patients with reconstruction, indicating its potential as a valuable tool for predicting cosmetic outcomes. However, this study primarily focused on the analysis of 3-dimensional simulation CT images and primarily concentrated on comparing cosmesis before and after radiation therapy, rather than directly evaluating cosmesis through high-quality visualizations.

To overcome these challenges, we put forth an innovative architecture named the Attention-Guided Denoising Diffusion Anomaly Detection (AG-DDAD), which harnesses both the high-quality generation capability of diffusion models and the salient feature discernment ability provided by DINO vision transformer (ViT) attention (Supplementary Information).
Our model can be trained in an unsupervised manner, utilizing readily available and unlabeled data from 1,237 patients with predominantly normal cosmesis (excellent to good), which account for the majority of patient population. This eliminates the need for expert annotation and manual delineation during evaluation. AG-DDAD provides direct and exceptional visualization outcomes by comparing the expected restoration results in cases of normal cosmesis, offering valuable insights into the factors contributing to poor cosmesis. To validate the performance of our model, we conducted experiments using a well-curated dataset comprising 300 patients who underwent breast-conserving surgery for breast cancer, alongside consensus labels obtained from clinical experts. The results demonstrated the superiority of our model over conventional rule-based methods and other state-of-the-art (SOTA) anomaly detection approaches.

\section{Related works}
\label{sec2}
\subsection{Denoising diffusion models}
Denoising diffusion models have garnered significant attention in the field of generative modeling \citep{croitoru2023diffusion}. These models aim to learn the underlying data distribution by iteratively denoising samples corrupted with Gaussian noise. The Denoising Diffusion Probabilistic Model (DDPM), a novel latent variable model inspired by nonequilibrium thermodynamics, was introduced by \cite{ho2020denoising}. It has been shown to generate high-quality image samples, rivaling the sample quality of generative adversarial networks (GANs) \citep{dhariwal2021diffusion}.

To address the slow sampling process observed in the pioneering work, \cite{song2020denoising} proposed a novel approach called Deterministic Diffusion Implicit Model (DDIM). DDIM leverages a diffusion model trained in a manner similar to the DDPM, harnessing non-Markovian processes. By employing this technique, DDIM introduces a deterministic framework with implicit modeling capabilities, resulting in a significant acceleration of the learning speed. Additionally, DDIM demonstrates the feasibility of deterministic inversion and interpolation of latent variables. These advancements not only enhance the efficiency of the diffusion model but also enable deterministic inversion and interpolation.

These diffusion models exhibit promising characteristics that can be effectively utilized in image editing and inpainting tasks. A notable development in this regard is Stochastic Differential Editing (SDEdit) \citep{meng2021sdedit}, which achieves a balance between realism and user input fidelity by performing the forward diffusion process only up to intermediate points, leveraging a diffusion model generative prior. This approach surpasses existing GAN-based methods across multiple tasks, as corroborated by perceptual studies conducted with human participants. Furthermore, RePaint presents a DDPM-based approach for free-form inpainting \citep{lugmayr2022repaint}, demonstrating its applicability to challenging masks. This method modifies reverse diffusion iterations specifically for unmasked regions without altering the original DDPM network, resulting in the generation of high-quality output images. These findings highlight the ability of diffusion models to produce high-quality images when dealing with inpainting tasks involving free-form masks.

\subsection{Unsupervised anomaly detection}
Advancements in anomaly detection have been significantly driven by unsupervised learning methods, which are trained on normal data and utilize scoring mechanisms to identify anomalies, or data points that deviate significantly from the norm \citep{chalapathy2019deep}. These techniques can be categorized into feature-based and reconstruction-based methods.

Feature-based methods employ self-supervised learning to acquire image features and detect anomalies by identifying differences in feature representations between normal and abnormal samples. Notable early work in this field includes DN2 \citep{bergman2020deep}, which utilized simple ResNets pretrained on ImageNet. Recent approaches have expanded upon this model. For example, PaDiM \citep{defard2021padim} employs locally constrained bag-of-features, SPADE \citep{cohen2020sub} utilizes a memory bank of extracted features, and PatchCore \citep{roth2022towards} incorporates a memory bank with neighborhood-aware patch-level features.

Reconstruction-based methods capitalize on the ability of models to accurately reconstruct normal data while exhibiting higher reconstruction errors for anomalies, using the extent of reconstruction error as an anomaly indicator. Early efforts utilized Variational Autoencoders \citep{kingma2013auto} to detect anomalies in skin disease images but were hindered by blurry reconstructions. The advent of GANs gave rise to models such as AnoGAN \citep{schlegl2019f} and Ganomaly \citep{akcay2019ganomaly}, which leveraged the increased reconstruction errors of GANs for abnormal data, surpassing prior SOTA models in anomaly detection. However, GANs faced difficulties with high-resolution images, poor reconstruction quality, and unstable training.

Recently, denoising diffusion models have gained increased attention for generating various data types due to their ability to produce high-quality images. This led to their application in anomaly detection. In the medical field, AnoDDPM demonstrated that diffusion models \citep{wyatt2022anoddpm}, when employing simplex noise instead of Gaussian noise, outperformed GANs in anomaly detection. Similarly, DiffusionAD \citep{zhang2023diffusionad}, which utilizes two sub-networks for denoising and segmentation, further exemplified the success of diffusion models in this domain. Concurrently, \cite{mousakhan2023anomaly} introduced DDAD, a novel method employing a target image-conditioned denoising process for reconstruction, thereby eliminating the need for sub-networks.

\subsection{Self-supervised vision transformer}
The Transformer model, introduced by \cite{vaswani2017attention}, has brought about a revolution in natural language processing through its attention mechanism and the absence of recurrent components, leading to remarkable advancements. Building upon this success, Transformers have recently been explored in image processing to harness their capability of capturing long-range dependencies in local image components. One noteworthy derivative is the vision transformer (ViT) model, which has demonstrated SOTA performance in image classification by replacing convolutions with multi-head self-attention mechanisms \citep{dosovitskiy2020image}. This development has sparked extensive research on ViT variants for various image processing tasks such as classification, object detection, and segmentation \citep{han2022survey}.

Moreover, recent findings suggest that ViT models benefit more from self-supervised learning compared to Convolutional Neural Networks. The flexibility of the self-attention mechanism enables ViT models to effectively capture semantic features in images. Notably, the distillation with no label (DINO) self-supervised learning \citep{caron2021emerging}, a label-free knowledge distillation method, has demonstrated its general semantic feature extraction capability, allowing it to serve as a task-agnostic feature extractor with SOTA performance for various downstream tasks.

Follow-up studies leveraging DINO-based ViT models have showcased the potential of these models as dense visual descriptors \citep{amir2021deep}. These features facilitate part-cosegmentation, providing highly localized semantic information with fine-grained spatial granularity. More recent advancements in DINO v2 have further enhanced the original DINO through augmented curated data and discriminative self-supervised pre-training methods, establishing it as a robust all-purpose pre-trained model with an accurate understanding of object parts \citep{oquab2023dinov2}. These ViT models with DINO-based self-supervised learning offer promising characteristics for attending to discriminative parts in images, making them effective for various vision tasks across different domains.

\section{Proposed framework}
\label{sec3}
\subsection{Overview of the proposed AG-DDAD model}
\begin{figure*}[!t]
\centering
\includegraphics[width=0.8\textwidth]{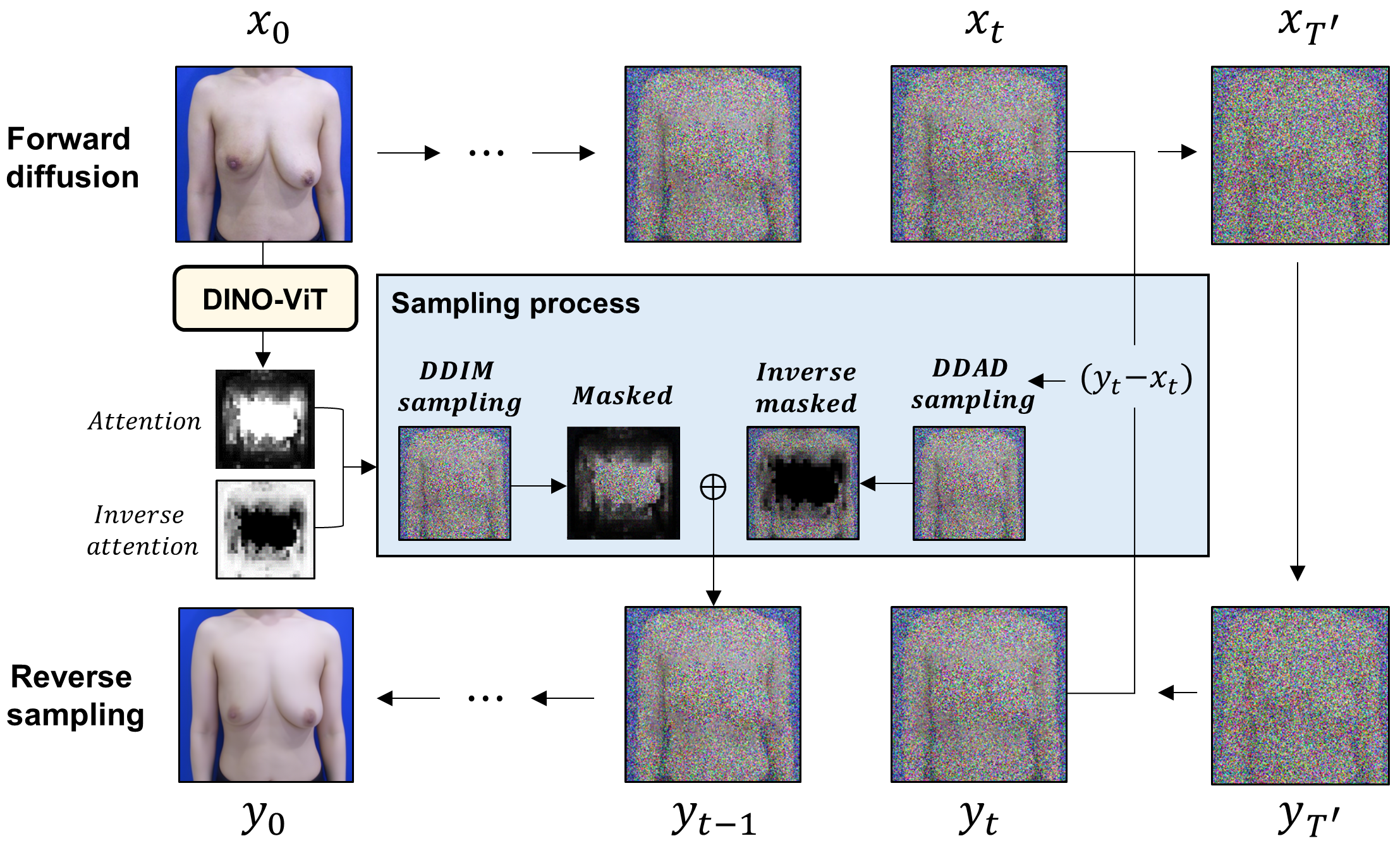}
\caption{The schematic illustration presents our proposed Attention-Guided Denoising Diffusion Anomaly Detection (AG-DDAD) model. This architecture harnesses the attention mechanism of the Vision Transformer (ViT), which has been self-trained via Distillation with no labels (DINO) methodology. This attention serves as a soft mask, fusing two disparate reverse sampling methods.} \label{fig:proposed_method}
\end{figure*}

The overall framework of the proposed AG-DDAD model is illustrated in Figure~\ref{fig:proposed_method}. The framework consists of two distinct stages designed to enhance the model's performance. The initial stage involves self-supervised learning of the ViT using the DINO method. In this stage, the primary objective is to learn a self-attention mask that distinguishes between discriminative and non-discriminative regions within an image. This mask serves as a crucial component for the subsequent stages of the model.


The second stage harnesses a sampling technique based on a diffusion model, guided by the soft attention mask derived from the initial stage. This stage selectively modifies the regions of interest within an image, aligning them with the distribution of the training data while preserving the visual appearance of the remaining areas. For this, the discriminative areas (e.g. breast) defined by the soft attention mask undergoes sampling process without any constraints forcing to keep original structure, while insignficant areas (e.g. background, body contour) remained unchanged with the constraint during the sampling process. 
The detailed processes of each stage are presented in the Methods section.

The detailed process of each stage is presented in the following sections.

\subsection{Self-supervised ViT for attention mask generation}
As suggested in previous studies, the adoption of DINO self-supervised learning empowers the ViT model to focus on semantically significant regions of images. This emphasis allows the model to pay attention to crucial areas that can also be utilized during subsequent image editing processes with the diffusion model.

Taking inspiration from prior work \citep{caron2021emerging}, we employed two identical networks based on the ViT model, referred to as the student and the teacher. The network architecture, denoted as $g$ and parameterized by $\theta$, consists of a backbone $f$ (preceding the final linear classifier) and a head $h$ for self-supervision.



\begin{align}
\label{Eq1}
g_{\theta} = h \circ f
\end{align}

Given an input image $x$, the model generates a prediction $P$ across $K$ dimensions. This prediction is obtained by normalizing the network output using the softmax function with a specified temperature $\tau$:
\begin{align}
\label{Eq2}
\quad P = g_{\theta}(x)
\end{align}
\begin{align}
\label{Eq3}
P = {\exp{(g_{\theta}(x)/\tau)} \over \Sigma_{k=1}^{K}{\exp{g_{\theta}(x)^{(k)}/\tau})}}
\end{align}
Here, the temperature parameter $\tau$ acts as a regulating factor that controls the sharpness of the output distribution.


To train the student model, parameterized by $\theta_{s}$, with the predictions of the teacher model, parameterized by $\theta_{t}$, we minimize the cross-entropy loss to match the distributions of their predictions $P_{s}$ and $P_{t}$:
\begin{align}
\label{Eq4}
\min_{\theta_{s}}-P_{t}\log{P_{s}}
\end{align}


The equation represented as Eq.~(\ref{Eq4}) can be adapted for self-supervised learning as follows. Initially, given an input image $x$, a global crop along with multiple local crops are obtained using the multi-crop strategy \citep{caron2021emerging} and random augmentations. This process aims to create various distorted views. Specifically, for a given image, a set $V = {x^{o}, x^{g}, x^{l}_1, ... x^{l}_L}$ is constructed. This set consists of one \emph{original} view $x^{o}$ without any augmentation, one \emph{global} view $x^{g}$, and $L$ \emph{local} views $x^{l}$ of smaller sizes.


The \emph{original} and \emph{global} views are passed through the teacher model, while all views are processed by the student model. This approach promotes a "global-local" correspondence, which is further enforced by the following optimization problem:
\begin{align}
\label{Eq5}
\mathcal{L} = \min_{\theta_{s}}{\sum_{x \in \{x^{o}, x^{g}\}}\sum_{\substack{x^{\prime} \in V \\ x^{\prime} \neq x}}~-g_{\theta_{t}}(x)\log{g_{\theta_{s}}(x^{\prime})}}
\end{align}

\begin{figure}[!t]
\centering
\includegraphics[width=0.5\textwidth]{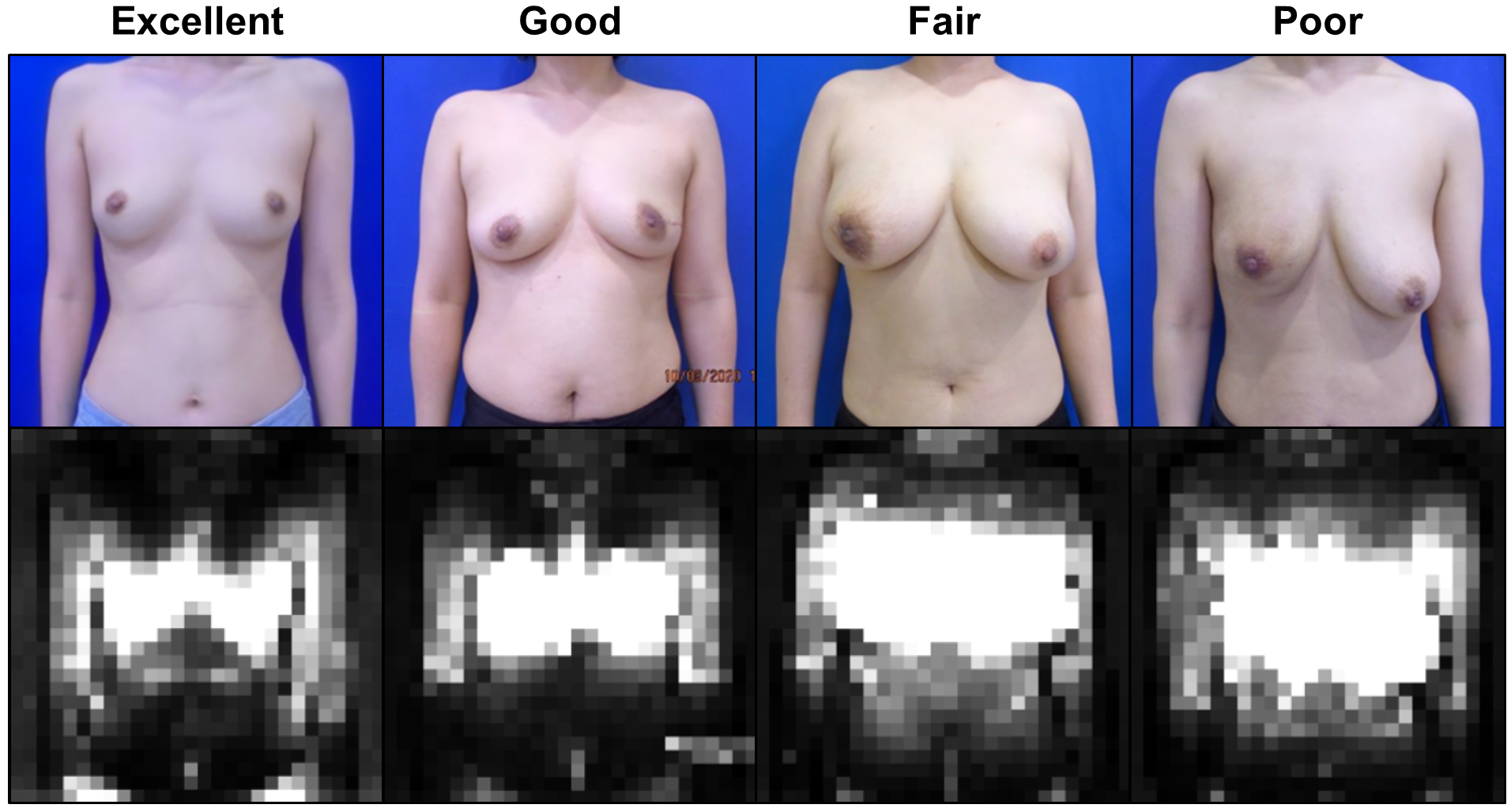}
\caption{Examples of the soft mask, derived from the attention weights of self-supervised Vision Transformer (ViT) models, each corresponding to different cosmesis groups.} \label{fig:attention}
\end{figure}

Upon employing the model trained in this manner, we noticed that the attention was concentrated mainly on the breast area in all cosmesis cases, ranging from excellent to poor, as depicted in Figure~\ref{fig:attention}. The attention weight matrix $A$ was then averaged across all attention heads, and the scale parameter $S$ was multiplied to obtain the soft mask $M$. This soft mask was subsequently utilized in the subsequent sampling process.


\begin{align}
\label{Eq6}
{M} = {S}{A}
\end{align}

\subsection{Anomaly scoring through high-quality reconstruction}

To train the diffusion model, we followed the approach proposed in the original DDPM work \citep{ho2020denoising}. Specifically, we ensured that the latent variable model $p_\theta(x_0) := \int{p_\theta(x_{0:T})~dx_{1:T}}$ generates latent variables $x_1, \ldots, x_T$ of the same dimension as the data $x_0 \sim q(x_0)$ by adding noise $\epsilon \sim \mathcal{N}(0,\textbf{I})$ at each time step $t$. In this way, $x_T$ follows a Gaussian distribution, $q(x_T) \sim \mathcal{N}(0,\textbf{I})$.

The forward process over $T$ steps, which gradually adds Gaussian noise following a predetermined variance schedule $\beta_{1}, \ldots, \beta_{T}$, can be represented by the fixed Markov chain, as shown in the following equation.



\begin{align}
\label{Eq7}
q(x_{1:T}|x_0) = \prod_{(t\geq1)}~q(x_{t}|x_{t-1})
\end{align}

\begin{align}
\label{Eq8}
q(x_{t}|x_{t-1}) = \mathcal{N}(x_{t};\sqrt{1 - \beta_{t}}x_{t-1},\beta_{t}\textbf{I})
\end{align}

A salient feature of the forward process is its ability to permit the sampling of ${x}_{t}$ at any given timestep $t$ in a closed form.

By employing the notation $\alpha_t := 1 - \beta_{t}$ and $\bar{\alpha}_{t} := \prod^{t}_{s=1}~\alpha_{s}$, it can be formulated that:

\begin{align}
\label{Eq9}
q(x_t|x_0) = \mathcal{N}(x_t;\sqrt{\bar{\alpha}_{t}}x_0,({1 - \bar{\alpha}_{t}})\textbf{I})
\end{align}

The reverse process, as opposed to simply adding noise, aims to restore the original data distribution. It is represented by $p_{\theta}(x_{t-1}|x_{t}) = \mathcal{N}(x_{t-1};\mu_{\theta}(x_t,t),\beta_{t}\textbf{I})$. In this formulation, the mean is derived from the learnable function $\epsilon^{(t)}_{\theta}(x_{t})$, and the learning objective involves minimizing $\vert \vert\epsilon^{(t)}_{\theta}(x_{t}) - \epsilon\vert \vert^{2}$.


In DDIM \citep{song2020denoising}, a deterministic non-Markovian sampling process is employed to accelerate the sampling procedure with minimal degradation in image quality. Specifically, DDIM introduces a forward process $q_{\sigma}(x_{t-1}|x_{t},x_{0})$ by defining a new variance schedule. Exploiting the unique property of the forward process, $x_{t}$ can be expressed as a linear combination of $x_{0}$ and a noise variable, denoted as $x_t = \sqrt{\alpha_{t}}x_{0} + \sqrt{1 - \alpha_{t}}\epsilon$.

Using this formulation, the denoised observation can be directly obtained from $x_{t}$ as follows:



\begin{align}
\label{Eq10}
f^{(t)}_{\theta} := (x_{t} - \sqrt{1 - \alpha_{t}}\cdot\epsilon^{(t)}_{\theta}(x_{t}))/\sqrt{\alpha_{t}}
\end{align}

Subsequently, we can define the generative process with a fixed prior 
$p_{\theta}(x_{T}) = \mathcal{N}(0,\textbf{I})$ and $p^{(t)}_{\theta}(x_{t-1}|x_{t}) = q_{\sigma}(x_{t-1}|x_{t},f_{\theta}^{(t)}(x_{t}))$.
As a result, $x_{t-1}$ can be expressed as:

\begin{align}
\label{Eq11}
x_{t-1} = \sqrt{\alpha_{t-1}}f^{(t)}_{\theta}(x_t) + \sqrt{1 - \alpha_{t-1} - \sigma_{t}^2}\cdot\epsilon^{(t)}_{\theta}(x_{t}) + \sigma_{t}\epsilon_{t}
\end{align}

Here, the stochasticity of the sampling process is determined by $\sigma_{t}$.

The connection between diffusion models and score matching \citep{song2019generative} was initially established by \cite{song2020score}, who introduced a score-based function to estimate the required deviation at each time step for generating a less noisy image. This can be expressed as follows:


\begin{align}
\label{Eq12}
\nabla_{x_{t}}\log p_{\theta}(x_{t}) = -{1 \over \sqrt{1 - \alpha}}\epsilon^{(t)}_{\theta}(x_{t})
\end{align}


This approach was later employed by \cite{dhariwal2021diffusion} to introduce conditional sampling for DDIM.


In the recent work on DDAD \citep{mousakhan2023anomaly}, the authors utilized this score-based function to introduce a conditional denoising process. Specifically, the authors devised a strategy to effectively conduct anomaly detection by introducing a condition to the unconditional score function. This strategy aimed to transform the data to be more akin to normal, while preserving its overall shape. The original image is $y_{0}$. For mathematical convenience, we set $y_t \sim q(x_t|x_0)$, but by using the same noise $\epsilon_\theta^{(t)}(x_t)$ that matches $x_t$. In other words, the deviation of $y_t$ from $x_t$ only exists in the signal component. The conditioned noised original image $y_t$ by introducing $\nabla_{x_{t}}\log p_{\theta}(x_{t}|y_{t})$, which can be formulated as follows with the Bayes' rule:

\begin{align}
\label{Eq13}
\nabla_{x_{t}}\log p_{\theta}(x_{t}|y_{t}) = \nabla_{x_{t}}\log p_{\theta}(x_{t}) + \nabla_{x_{t}}\log p_{\theta}(y_{t}|x_{t})
\end{align}

To penalize deviations from $x_t$, a divergence by $y_t - x_t$ was calculated, from the assumption of matched noise $\epsilon_\theta^{(t)}$. An adjusted noise term was defined as:

\begin{align}
\label{Eq14}
\hat{\epsilon} = \epsilon^{(t)}_{\theta}(x_{t}) - \omega\sqrt{1 - \alpha_{t}}(y_t - x_t)
\end{align}


As a result, the sampling process in DDAD can be represented as follows. In comparison to DDIM sampling, it not only aims to restore the sample closer to the learned distribution during reverse sampling but also penalizes deviations that affect the preservation of the original shape:

\begin{align}
\label{Eq15}
x_{t-1} = \sqrt{\alpha_{t-1}}\hat{f}^{(t)}_{\theta}(x_t) + \sqrt{1 - \alpha_{t-1} - \sigma^{2}_{t}}\hat{\epsilon} + \sigma_{t}\epsilon_{t}
\end{align}

However, as indicated in our ablation study, directly applying DDAD to breast cancer patient data brought about significant limitations in terms of shape transformation. These limitations may stem from the noise correction process in Eq.~\ref{Eq14}, where the expectation of noise deviates from the Gaussian distribution, making it challenging to effectively restore normal cosmesis seen during training.



Conversely, when DDIM sampling was applied without such constraints, the noise followed a Gaussian distribution, resulting in the generated data transforming into images resembling normal cosmesis. However, it should be noted that this transformation also led to significant changes in the overall image details.

Therefore, we employed a hybrid approach that integrates the sampling methods of DDIM and DDAD. This hybrid approach allows us to selectively apply the appropriate sampling method based on the specific regions that require transformation for anomaly detection (DDIM sampling) and those that need to be preserved (DDAD sampling).


To achieve this, we utilized a soft mask $M$ obtained from Eq.~\ref{Eq6}, which indicates the regions that require transformation. Additionally, we derived an inverse soft mask $1 - M$, which represents the areas that need to be preserved.


\begin{align}
\label{Eq16}
\tilde{\epsilon} = M \circ \epsilon + (1 - M) \circ \hat{\epsilon}
\end{align}

Plugging Eq.~\ref{Eq16} into Eq.~\ref{Eq11} enables to selectively transform the discriminative components of image while conserving the integrity of remaining parts. This process occurs seamlessly due to the connection facilitated by the soft mask, which possesses a continuous range of values. 

Analytically, Eq.~\ref{Eq16} can also be interpreted as rectifying $\hat\epsilon$ to maintain its proximity to the Gaussian annulus as it integrates with pure noise. As a result, less error will be accumulated. This could potentially be another explanation as to why the suggested approach shows superior performance compared to the naive DDAD.


\begin{table}[!t]
  \centering
  \caption{Details of datasets and cosmesis groups}
        \resizebox{0.48\textwidth}{!}{
    \begin{tabular}{ccccccc}
    \toprule
    \multirow{2}[4]{*}{\textbf{Dataset}} & \multirow{2}[4]{*}{\textbf{Unlabeled}} & \multicolumn{4}{c}{\textbf{Labeled}} & \multirow{2}[4]{*}{\textbf{Total}} \\
\cmidrule{3-6}          &       & \textbf{Excellent} & \textbf{Good} & \textbf{Fair} & \textbf{Poor} &  \\
    \midrule
    \textbf{Train} & 1,237 & -     & -     & -     & -     & 1,237 \\
    \textbf{Test} & -     & 74    & 99    & 75    & 52    & 300 \\
    \bottomrule
    \end{tabular}%
    }
  \label{tab:data}%
\end{table}%

\section{Implementation details}
In this section, we provide the implementation details of our experiments. All experiments were conducted on a system equipped with two GeForce RTX 3090 graphics cards. We used PyTorch 1.12.1 for model development and training. Performance evaluation and statistical analysis were performed using the Python library sklearn 1.1.3 and R-software 4.3.1, respectively.


\label{sec4}
\subsection{Details of dataset}
Table~\ref{tab:data} provides a comprehensive overview of the data distribution for each cosmesis group. For our experiments, we utilized patient data from the Yonsei Cancer Center, involving individuals who underwent breast-conserving surgery for breast cancer with curative intent between June 2018 and June 2020. Specifically, we employed unlabeled photographic data from a total of 1,237 patients predominantly with normal cosmesis (excellent to good), which account for the majority of patient proportion, to train the ViT model for attention mask generation and the diffusion model for anomaly scoring.

To evaluate the performance of our proposed approach, we created a separate dataset consisting of 300 patients, ensuring that there was no overlap with the training data. The photographs in the evaluation dataset were independently labeled by two board-certified radiation oncologists using the modified Harvard-Harris cosmetic scale. The labelers were blinded to each other's annotations, and any disagreements were resolved through consensus. The consensus labels obtained from the experts were utilized as the ground truth for performance evaluation.



\subsection{Details of DINO self-supervised ViT for attention guidance}
In order to generate soft attention masks, we employed the ViT-s/8 model for both the teacher and student models in the DINO self-supervised learning process. The input photographs were resized to 224 $\times$ 224 and subsequently normalized. The DINO self-supervised learning was conducted using a range of global crop scales from 0.4 to 1.0, and local crop scales from 0.05 to 0.40. To train the model, we utilized a multi-crop strategy which involved incorporating two global crops and eight local crops. The training process consisted of 500 epochs, utilizing the entire training data, with a learning rate of 0.00001. We employed the AdamW optimizer \citep{loshchilov2017decoupled} and maintained a batch size of 16. Considering that the ViT-s/8 model has six attention heads, we computed the average attention from these six heads to obtain the attention mask. The scale parameter $S$ was set to 500.


\subsection{Details of diffusion model training and sampling}
In our approach, we employed a modified UNet architecture, as described in \cite{song2020score}, as our denoising model. Each photograph underwent meticulous resizing to a resolution of 256 $\times$ 256 and subsequent normalization, serving as input for our model. The training phase involved a rigorous regimen spanning 300,000 steps, utilizing the Adam optimizer with a learning rate of 0.0001 and a batch size of 4.

Inspired by previous studies that demonstrated different effects of time step manipulation during reverse sampling, we adopted a forward step approach based on \citep{meng2021sdedit}, where increasing the time step $t$ to a maximum value $T$ results in random image generation, while restricting it to a midpoint ${T\prime}$ preserves the approximate shape with potential deformation. In our implementation, we incorporated noise levels up to ${T\prime} = 500$ out of a total time step $T = 1000$, a value determined empirically. To expedite the sampling process without significant sacrifice in generation quality, we rescaled the entire time step from $T = 1000$ to a more manageable $T = 250$ for our experimental procedures.



\subsection{Details of evaluation}
We evaluated the performance of our model through three distinct approaches. Firstly, we conducted an evaluation of the anomaly scores across different cosmesis groups. We examined the anomaly scores of patients within each cosmesis group and verified whether there was a significant difference in anomaly scores according to the cosmesis. 
Secondly, we assessed the performance of the proposed AG-DDAD model as an anomaly detection model. Specifically, we investigated its ability to identify subjects with poor cosmesis from a mixed group comprising both excellent and poor cosmesis cases. We measured the model's performance using the area under the curve (AUC) metric, a commonly used measure for evaluating anomaly detection models. To ensure a fair comparison, we calculated the anomaly scores for each SOTA anomaly detection method as suggested in their respective original papers. For diffusion model-based methods, we computed anomaly scores based on the reconstruction error, specifically the mean squared error, to fairly compare the reconstruction quality of the diffusion model. 
Lastly, we compared our method as an objective approach with the widely used BCCT.core program, which is a rule-based, non-automated breast cosmesis evaluation method after surgery that does not employ AI. BCCT.core program calculates cosmesis based on the relative positions of the breast contour, nipple, scale marker, and sternal notch, which are manually defined by the user. We evaluated the degree of agreement of our method when compared with expert labels against this method. The degree of agreement was assessed using the Cohen's kappa coefficient.

The independent samples t-test was utilized to assess the disparities in anomaly scores between the cosmesis groups. The DeLong's test \citep{delong1988comparing} was employed to examine the statistical significance of variations in detection accuracy among the different anomaly detection methods employed in the study. All statistical tests performed in this analysis were two-tailed, thereby considering both positive and negative differences in relation to the null hypothesis. For all analyses, P-values $< 0.05$ and the non-overlapping confidence intervals were defined as statistically significant differences.

\section{Experimental results}
\label{sec6}
\subsection{Differences of anomaly scores between cosmesis groups}
\begin{figure}[!t]
\centering
\includegraphics[width=0.5\textwidth]{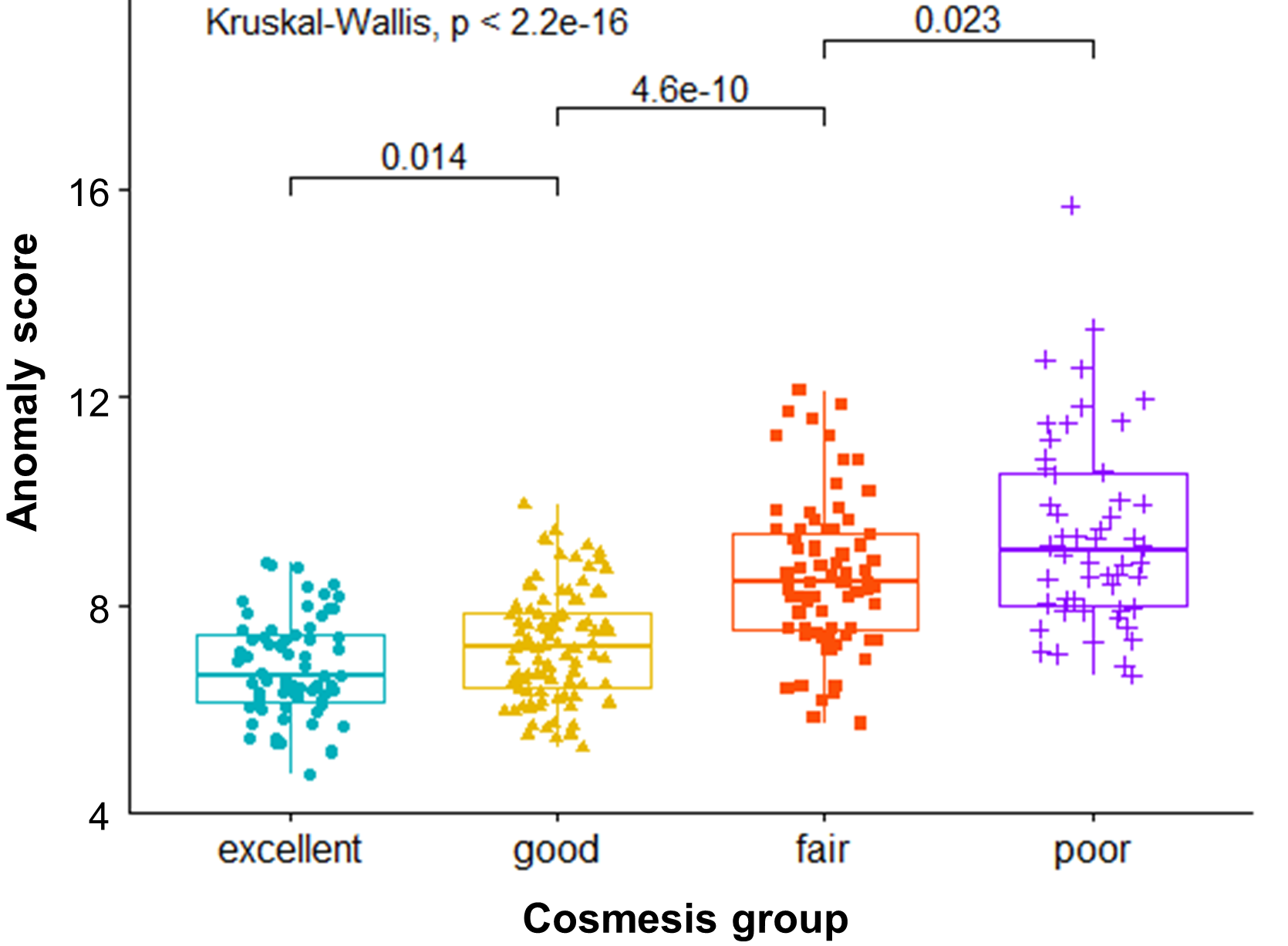}
\caption{Comparative analysis of anomaly scores across different cosmesis groups. There exists a statistically significant escalation in the anomaly score concurrent with the deterioration of cosmesis, which is observed across all groups.} \label{fig:groups}
\end{figure}

Figure~\ref{fig:groups} presents the anomaly scores estimated by our proposed AG-DDAD model for each cosmesis group. The graph clearly demonstrates a consistent pattern of increasing anomaly scores as the cosmesis quality declines. The median scores (with interquartile ranges) for the excellent, good, fair, and poor groups were as follows: median 6.628 (interquartile range [IQR], 6.119-7.412) for the excellent group, median 7.175 (IQR, 6.412-7.839) for the good group, median 8.427 (IQR, 7.534-9.383) for the fair group, and median 9.052 (IQR, 8.001-10.520) for the poor group. Significantly, there were notable differences in scores observed between these groups.


Interestingly, the disparity in scores between the groups with good cosmesis quality (excellent versus good) was relatively minor. However, a substantial distinction was observed between the good and poor cosmesis groups (excellent, good vs. fair, poor). These findings suggest that our model exhibits promising potential as an anomaly detection tool, specifically for identifying cases of poor cosmesis.


\begin{table}[!t]
  \centering
  \caption{Performance comparison with SOTA anomaly detection models}
  \begin{adjustbox}{width=0.5\textwidth}
    \begin{tabular}{cc}
    \toprule
    \textbf{Methods} & \textbf{AUC (95\% CI)} \\
    \midrule
    \textbf{\textit{Feature-based}} &  \\
     PaDiM \citep{defard2021padim} & 0.556 (0.448 - 0.664) \\
     DFM \citep{wang2021dynamic} & 0.494 (0.462 - 0.527) \\
     STFPM \citep{wang2021student} & 0.685 (0.589 - 0.780) \\
     FastFlow \citep{yu2021fastflow} & 0.586 (0.481 - 0.691) \\
     CSFlow \citep{rudolph2022fully} & 0.644 (0.546 - 0.743) \\
     CFA \citep{lee2022cfa} & 0.511 (0.399 - 0.624) \\
     PatchCore \citep{roth2022towards} & 0.710 (0.619 - 0.800) \\
    Reverse Distillation \citep{deng2022anomaly} & 0.652 (0.552 - 0.751) \\
        \midrule
    \textbf{\textit{Reconstruction-based}} &  \\
     Ganomaly \citep{akcay2019ganomaly} & 0.730 (0.642 - 0.817) \\
     DRAEM \citep{zavrtanik2021draem} & 0.676 (0.581 - 0.770) \\
     DDPM \citep{ho2020denoising} & 0.815 (0.742 - 0.887) \\
     DDIM \citep{song2020denoising} & 0.866 (0.804 - 0.928) \\
     DDAD-10 \citep{mousakhan2023anomaly} & 0.859 (0.794 - 0.925) \\
     \textbf{Propsoed (AG-DDAD)} & \textbf{0.921 (0.876 - 0.965)} \\
    \bottomrule
    \end{tabular}%
    \end{adjustbox}
  \label{tab:anomaly_detection}%
\end{table}%

\begin{figure*}[!t]
\centering
\includegraphics[width=1.0\textwidth]{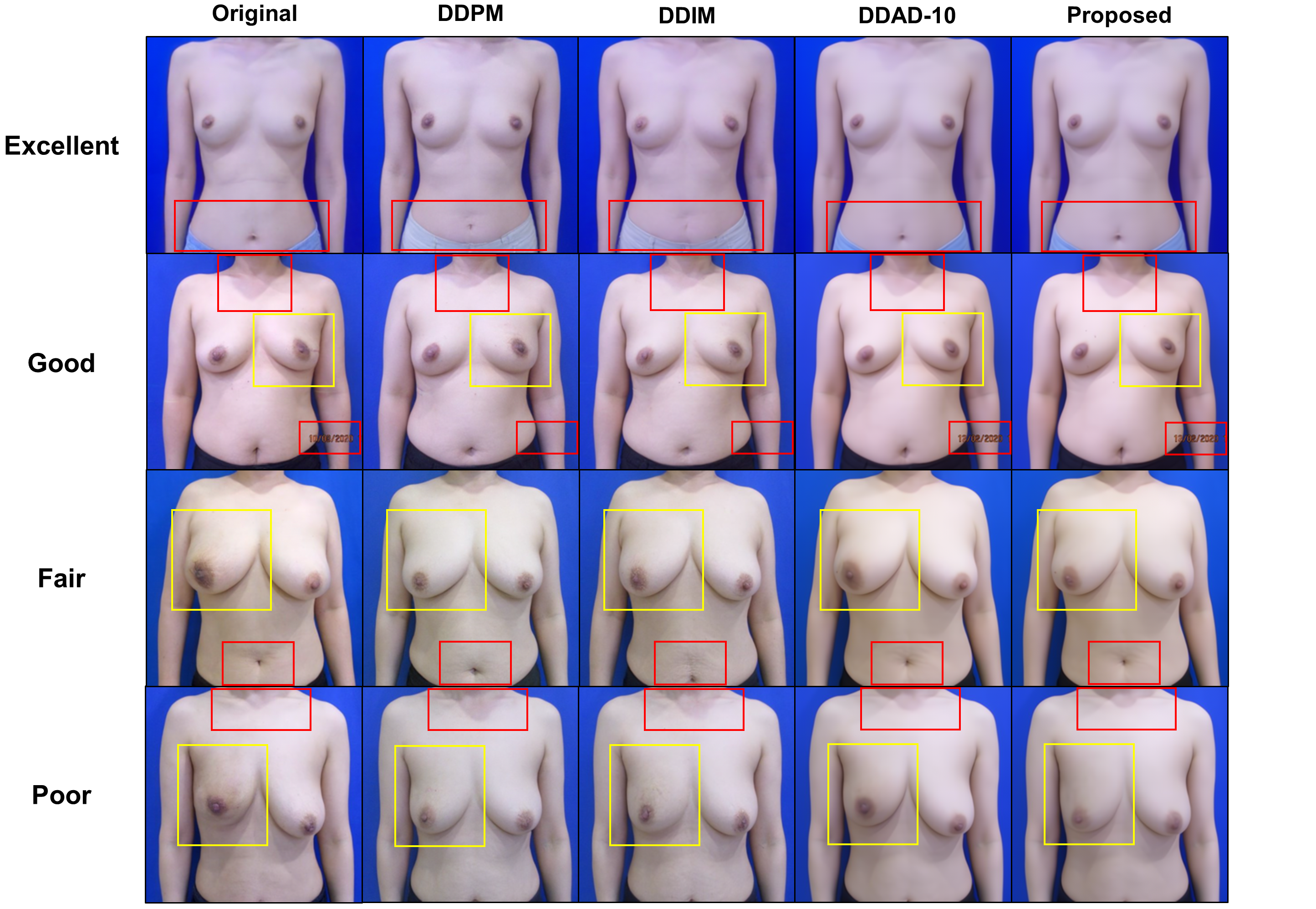}
\caption{Comparison of reconstruction outcomes for normal cosmesis between the proposed methodologies and other diffusion model-based methods. Other diffusion model-based approaches either distort details of the image that should be preserved (red box) or fail to sufficiently transform areas that need modification for achieving normal cosmesis (yellow box). Conversely, the proposed framework effectively transforms only the requisite portions while preserving the rest, as illustrated.} \label{fig:comparison}
\end{figure*}

\subsection{Performance comparison with anomaly detection models}
Subsequently, we proceeded to evaluate the effectiveness of our proposed model in the context of anomaly detection. Specifically, we examined how accurately a model trained unlabeled data containing predominantly normal cosmesis could identify poor cosmesis images using anomaly scores. To assess its performance, we compared our model against SOTA feature-based and reconstruction-based anomaly detection models.

Table~\ref{tab:anomaly_detection} presents the performances of various anomaly detection models, including our proposed approach. Overall, the reconstruction-based anomaly detection models demonstrated superior performance compared to the feature-based models. Among the reconstruction-based methods, the diffusion model-based techniques exhibited the highest level of performance. Notably, our model outperformed all others, a result that was statistically significant.



Figure~\ref{fig:comparison} provides a qualitative comparison of the reconstruction results obtained from diffusion model-based anomaly detection models and our proposed model. As depicted in the figure, while other models struggled with issues such as failing to correctly modify parts that should be changed (yellow box) or changing parts that should remain unchanged (red box), our model accurately modified only the necessary parts (areas with poor cosmesis) while preserving the rest.

\begin{table}[!t]
  \centering
  \caption{Comparison of agreement with conventional rule-based method}
    \begin{tabular}{cc}
    \toprule
    \textbf{Methods} & \textbf{Cohen's kappa (95\% CI)} \\
    \midrule
    BCCT.core program & 0.67 (0.61-0.73) \\
    Proposed & 0.60 (0.53-0.67) \\
    \bottomrule
    \end{tabular}%
  \label{tab:agreement}%
\end{table}%

In particular, the naive DDPM or DDIM models tended to transform the patient's image into one with overall good cosmesis, but at the expense of significant changes in detail. Examples of such changes include altering the color of the patient's pants in the output image, removing a marker in the good image, shifting the position of the navel in the fair image, or modifying the shape of the shadow in the neck area in the poor image. These unnecessary alterations compromised the anomaly detection performance of these models.

In contrast, the DDAD model, with its regularization effect that discourages deviations from the original image, introduced minimal unnecessary changes. However, this constraint resulted in the model's limited ability to sufficiently transform the fair and poor images to resemble excellent cosmesis. On the other hand, our proposed AG-DDAD model accurately targeted only the necessary areas for modification while preserving the parts that should remain unchanged. Consequently, it successfully avoided unnecessary alterations and achieved the highest performance in anomaly detection.




\begin{figure*}[!t]
\centering
\includegraphics[width=0.77\textwidth]{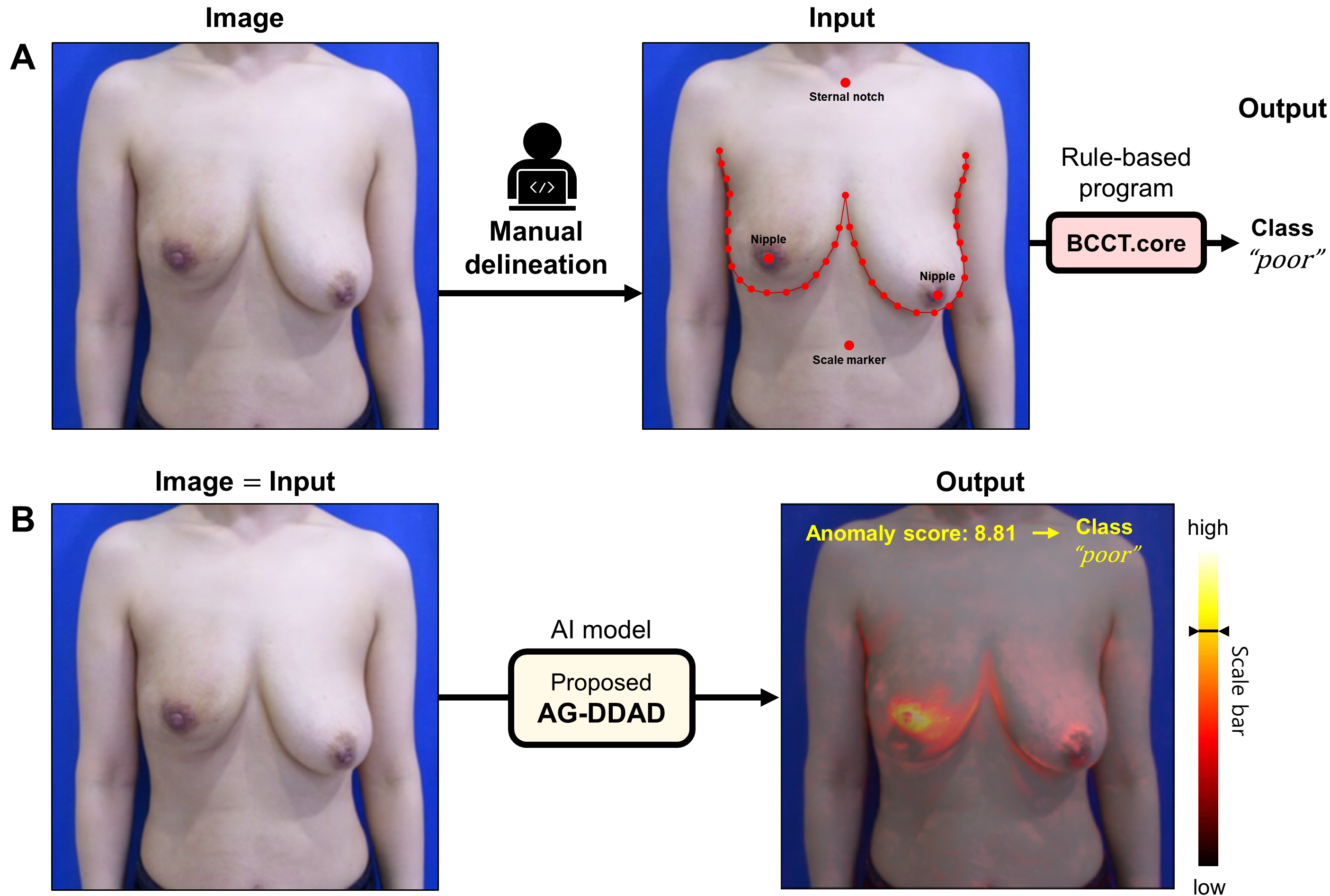}
\caption{Comparison between the conventional rule-based scoring method, the BCCT.core program and the proposed framework. (A) The BCCT.core program necessitates manual delineation of the breast contour and anatomical markers (red dots) to produce class results, yet it fails to provide any quantifiable scores or visualization outcomes. (B) The proposed framework utilizing the AG-DDAD model, on the other hand, generates quantifiable anomaly scores conducive to comparison, solely based on the image itself, without requiring any additional processes. This framework also facilitates the visualization of corresponding anomaly maps.
} \label{fig:bcct}
\end{figure*}

\subsection{Performance comparison with rule-based scoring method}
We compared our model with the BCCT.core program, a conventional rule-based scoring method. The agreement between the BCCT.core program and the consensus label of experts, as measured by Cohen's kappa value, was found to be 0.67 (0.50-0.72) (Table~\ref{tab:agreement}). Our model demonstrated a slightly lower but comparable kappa value of 0.60 (0.53-0.67), as indicated by the overlapping confidence intervals. Considering the relatively modest level of agreement among experts, these results suggest that our model holds clinical value as an objective tool for breast cosmesis evaluation.

Our model offers significant advantages over traditional rule-based scoring systems like the BCCT.core program. The BCCT.core program requires manual delineation of markers such as the position of the nipple, breast contour, scale marker, and sternal notch to accurately represent the modified Harvard-Harris cosmetic scale. It then calculates numerical relationships between these elements to predict cosmesis in four groups. In contrast, our model can provide a cosmesis prediction using just a single photograph, without requiring any manual delineation process. It also generates a high-quality visualization of the predicted good cosmesis and presents an anomaly map that visualizes the differences from the original image. Furthermore, instead of providing only four classes, our model provides a quantifiable score that indicates the degree of anomaly, enabling the evaluation of cosmesis as a continuous variable (Figure~\ref{fig:bcct}).

\subsection{Ablation study}
In this section, we conducted an ablation study to evaluate the individual contributions of the key methods incorporated in our proposed AG-DDAD model.

\subsubsection{Attention guidance}
\begin{figure*}[!t]
\centering
\includegraphics[width=1.0\textwidth]{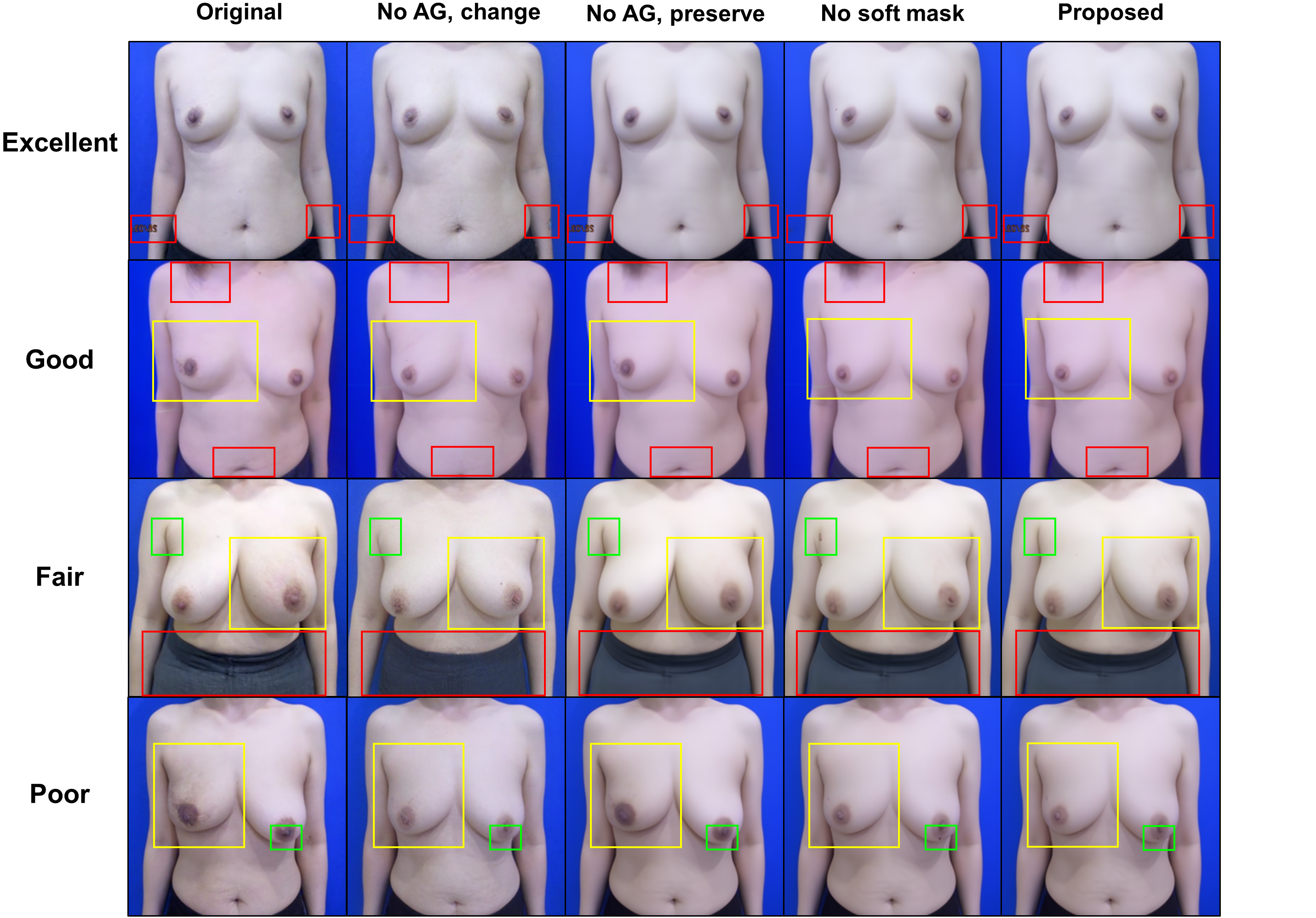}
\caption{
Comparison of reconstruction results for normal cosmesis in ablation studies of key components. When reverse sampling was applied without attention guidance (AG), either all areas underwent transformation, distorting even the details that should be preserved (red box), or all areas were preserved, failing to sufficiently transform to achieve normal cosmesis (yellow box). When a hard mask was employed instead of a soft mask, an unnatural artifact was observed due to the lack of smooth blending at the boundary (green box).
} \label{fig:ablation}
\end{figure*}

First, we investigated the impact of removing the attention guidance. When the attention guidance was excluded, the sampling in the reverse process was either performed solely using the DDIM sampling method or with a fixed high $\omega$ value using the DDAD sampling method. The results of this ablation are presented in Table~\ref{tab:ablation} and Figure~\ref{fig:ablation}.

Without attention guidance, both methods experienced a decline in performance. This decrease in performance can be attributed to two factors. In the case of DDIM sampling, the model erroneously modified unnecessary parts of the entire image (red box). On the other hand, when using the high $\omega$ value DDAD sampling method, the model failed to adequately modify the anomalous parts (yellow box). These findings emphasize the significance of attention guidance in our model for attaining optimal performance.



\subsubsection{soft mask for mixed sampling}
In our model, we incorporated a soft mask for mixed sampling, which combines the DDIM sampling and high $\omega$ value DDAD sampling methods during the sampling process. We also considered using a hard mask, obtained by thresholding the attention mask, as seen in previous studies utilizing diffusion model-based anomaly detection models \citep{bercea2023mask}. However, adopting a hard mask resulted in a performance decrease. Moreover, qualitative analysis revealed the presence of unnatural artifacts at the boundaries of the mask in the image (green box). These observations underscore the importance of employing a soft mask in our model to preserve optimal performance and image quality. The use of a soft mask enables a smoother transition at the mask boundaries, thereby preventing the introduction of artifacts and ensuring a more natural-looking result.


\begin{table}[!t]
  \centering
  \caption{Ablation studies of key components}
    \begin{tabular}{cc}
    \toprule
    \textbf{Methods} & \textbf{AUC (95\% CI)} \\
    \midrule
    \textbf{Proposed} & \textbf{0.921 (0.876 - 0.965)} \\
    w/o attention guidance (change) & 0.831 (0.758 - 0.904) \\
    w/o attention guidance (preserve) & 0.819 (0.744 - 0.894) \\
    w/o soft mask & 0.865 (0.803 - 0.927) \\
    \bottomrule
    \end{tabular}%
  \label{tab:ablation}%
\end{table}%

\section{Discussion and Conclusion}
\label{sec7}
The remarkable performance of deep learning-based AI models across diverse domains has spurred significant interest in applying AI to the field of medicine, yielding impressive results \citep{wang2019deep}. However, prevailing AI models in the medical domain heavily rely on conventional supervised learning methods that necessitate expert labels. This reliance poses a substantial limitation, particularly in medicine, where acquiring a substantial quantity of expert labels proves challenging and costly \citep{krishnan2022self}. 
Even when an abundance of expert label data is available, the tasks involving subjective judgments, like breast cosmesis evaluation in this work, introduce the risk of model learning biases from the labeler's subjective assessments.
Despite efforts in the medical domain to address this label dependency issue through unsupervised learning \citep{raza2021tour}, self-supervised learning \citep{krishnan2022self}, and semi-supervised learning \citep{chebli2018semi}, these approaches have encountered limited success within confined domains.

In order to address these limitations, we propose a novel automated approach, AG-DDAD model, for cosmesis evaluation following breast cancer surgery, which has demonstrated significant potential and holds great promise in clinical application. Specifically, we effectively integrate the property of the DINO self-supervised ViT that attends to important regions in images, which underlines the general semantic feature extraction capability, with the diffusion model that enables high-quality reconstruction. 
To achieve this, we employ a two-stage approach, where we first generate a soft mask using ViT attention and incorporate it into the reverse sampling process of the diffusion model. This ensures that the discriminative regions within image between normal and abnormal subjects with DDIM sampling, while preserving other shape with highly conditioned DDAD sampling. As a result, only the necessary regions undergo transformation into normal images, effectively capturing the desired changes. 

We approached the problem of breast cosmesis evaluation from an unsupervised anomaly detection perspective by training a diffusion model using excellent to good cosmesis data as normal samples. The effectiveness of our method was validated using real-world data, demonstrating its potential clinical applicability. \add{Through the AG-DDAD model, we successfully generated high-quality visualizations for predicting normal cosmesis outcomes on a per-patient basis.}
Furthermore, we computed quantifiable scores based on reconstruction errors and calculated group-level anomaly scores, revealing statistically significant differences in anomaly scores as cosmesis worsened. In comparison to widely adopted rule-based programs for objective cosmesis evaluation, our method offers a fully automated evaluation framework that eliminates the need for laborious delineation of breast contours or identification of anatomical markers. Unlike the BCCT.core program, which provides results for only four cosmesis classes, our method provides quantifiable scores and anomaly maps, enabling the estimation of the degree of cosmesis deterioration even within the same group. This allows for an examination of the underlying evidence used by the model to make judgments, thereby verifying their appropriateness (Figure~\ref{fig:bcct}).
In terms of clinical application, our method can be easily implemented as it can be trained with a predominantly normal cosmesis database in an unsupervised manner, without requiring a large number of data points for each cosmesis class or ground truth labels. Moreover, our approach presents the potential to provide an automated and objective evaluation method for tasks that previously relied heavily on subjective judgments by experts, thereby mitigating the risk of labeler bias inherent in conventional supervised learning approaches. 
\add{Lastly, the realistic and high-quality prediction of normal cosmesis outcomes for individual patients facilitates both retrospective evaluation of treatment success and prospective prediction of expected outcomes. This feature holds valuable implications for both patients and physicians in terms of treatment decision-making.}

From the perspective of the anomaly detection model itself, our model demonstrated SOTA performance, significantly surpassing existing anomaly detection models. Previous research has explored the use of diffusion models for anomaly detection by leveraging their ability to reconstruct images to match the learned distribution while preserving the overall structure, as demonstrated in works such as SDEdit \citep{meng2021sdedit}. However, a common limitation of these studies is the inherent stochasticity of the model, resulting in large reconstruction errors even in normal regions of the image, beyond the anomalous regions.
In the recent work that proposed DDAD \citep{mousakhan2023anomaly} attempts were made to address this issue by penalizing deviations from the original image. However, in our experiments, penalizing the entire image using DDAD had limitations in effectively transforming the image into normal cosmesis, leading to suboptimal performance. Therefore, we employed soft masks obtained from the ViT attention mechanism to selectively apply different sampling approaches to specific regions of the image. This allowed us to transform only the discriminative parts without requiring explicit supervision.
The comparison with existing diffusion model-based anomaly detection models and the results of our ablation study clearly demonstrate the effectiveness of our proposed approach, supporting the potential application of our method beyond breast cosmesis evaluation, extending to diverse anomaly detection tasks.

Our study has several limitations. Firstly, when comparing the agreement with ground truth labels between our method and BCCT.core, we found that BCCT.core program exhibited higher agreement, although not statistically significant. The higher agreement between BCCT.core and expert assessments is expected since the cosmesis measurement scale in BCCT.core directly reflects the modified Harvard-Harris cosmetic scale. The predefined rule-based approach of BCCT.core program, designed to align with the scoring system, may contribute to higher scores. Nevertheless, the overlapping confidence intervals suggest that the performance of the two methods can be considered comparable.
Secondly, our method requires slightly more time per image compared to conventional classifier models. The evaluation of cosmesis for a single patient using our approach takes approximately 15 seconds, whereas simple classification models can yield results within 1 second \citep{canziani2016analysis}. To address this limitation, future studies may explore the incorporation of acceleration techniques proposed for diffusion models.
Lastly, while we evaluated model performance using an evaluation set with expert consensus labeling from a distinct patient population, external validation using images captured in different environments and with diverse camera settings has not been conducted. Thus, further investigations at different institutions are warranted.

Nevertheless, our research presents a novel automated approach, AG-DDAD, for breast cosmesis evaluation, which shows promising potential for clinical application. By addressing the limitations of conventional supervised learning and existing anomaly detection models, our method offers a more objective and efficient evaluation framework. The integration of the DINO self-supervised ViT and diffusion model allows for robust anomaly detection and the quantification of cosmesis quality. Further investigations and improvements are warranted to fully exploit the capabilities of our approach and facilitate its wider adoption in clinical practice.

\section{Acknowledgement}
\label{sec7}
\add{This study was supported by a new faculty research seed money grant of Yonsei University College of Medicine for 2023 (2023-32-0068).}

\setcounter{figure}{0}
\setcounter{table}{0}
\renewcommand{\thefigure}{A\arabic{figure}}

\setcounter{section}{0}
\renewcommand*{\thesection}{Appendix \Alph{section}}
\begin{appendix}
\section{Ethic committee approval}
\label{sec_A}
The patient photograph data used in this study were obtained with the ethical approval of the Institutional Review Board (IRB) of Severance Hospital, under the IRB number 4-2022-0176.

\section{Modified Harvard-Harris cosmetic scale}
The Harvard cosmetic scale was proposed as a means to assess the cosmetic outcome following breast cancer surgery \citep{harris1979analysis}. This scale system categorizes the aesthetic results into four levels: "excellent," "good," "fair," and "poor." The evaluation of aesthetics within this system considers a comprehensive range of factors, including breast symmetry, nipple position, skin color, and the scars.

To enhance the inter-observer agreement and reproducibility of the scale, the Modified Harvard-Harris cosmetic scale was introduced \citep{zellars2009feasibility}. This modified version provides a more detailed description for each category, aiming to mitigate the brevity of the original scale. The inclusion of additional descriptive information can improve the consistency and reliability of evaluations. Refer to Table~\ref{tab:harvard} for the detailed descriptions associated with each cosmesis category.

\begin{table}[!t]
  \centering
  \caption{Modified Harvard-Harris cosmetic Scale}
\begin{adjustbox}{width=0.5\textwidth}
    \begin{tabular}{cp{24em}}
    \toprule
    \textbf{Cosmesis} & \multicolumn{1}{c}{\textbf{Description}} \\
    \midrule
    \textbf{Excellent} & When comared with the baseline image, there is minimal or no difference in size or shape or consistency of the breast. There may be mild thickening or scar tissue within the breast or skin, but not enough to change the appearance. \\
        \midrule
    \textbf{Good} & When compared with the baseline image, there is mild asymmetry or slight difference in the size or shape of the breast. Mild reddening or darkening of the breast. The thickening or scar tissue within the breast causes only a mild change in the shape. \\
        \midrule
    \textbf{Fair} & When compared with the baseline image, there is moderate deformity with obvious difference in the size and shape of breast. This change involves one quarter or less of the breast. There is moderate thickening or scar tissue of the skin and the breast and obvious color change. \\
        \midrule
    \textbf{Poor} & When compared with the baseline image, there is marked change in the appearance of the breast involving more than one quarter of the breast tissue. The skin change sare very obvious. There is severe scarring and thickening of the breast. In retrospect, mastectomy would have been a better option \\
    \bottomrule
    \end{tabular}%
\end{adjustbox}
  \label{tab:harvard}%
\end{table}%

\end{appendix}

\bibliographystyle{model2-names.bst}\biboptions{authoryear}
\bibliography{refs}

\end{document}